\documentclass[sigconf]{acmart}

\usepackage{algorithm}
\usepackage{algorithmic}

\AtBeginDocument{%
  \providecommand\BibTeX{{%
    \normalfont B\kern-0.5em{\scshape i\kern-0.25em b}\kern-0.8em\TeX}}}

\begin{document}

\title[PGAHum]{PGAHum: Prior-Guided Geometry and Appearance Learning for High-Fidelity Animatable Human Reconstruction}

\author{Hao Wang}
\authornotemark[1]
\affiliation{%
  \institution{Jilin University}
  \city{Changchun}
  \country{China}
  \postcode{130000}
}
\email{whao22@mails.jlu.edu.cn}

\author{Qingshan Xu}
\authornote{Both authors contributed equally to this research.}
\affiliation{%
  \institution{Nanyang Technological University}
  \country{Singapore}
}
\email{qingshan.xu@ntu.edu.sg}

\author{Hongyuan Chen}
\affiliation{%
  \institution{Jilin University}
  \streetaddress{Qianjin Street}
  \city{Changchun}
  \country{China}
}
\email{chenhy5521@mails.jlu.edu.cn}

\author{Rui Ma}
\authornote{Corresponding author.}
\affiliation{%
  \institution{Jilin University}
  \institution{Engineering Research Center of Knowledge-Driven Human-Machine Intelligence, MOE, China} 
  \country{}
}
\email{ruim@jlu.edu.cn}

\renewcommand{\shortauthors}{Hao et al.}


\settopmatter{printacmref=false} 

\renewcommand\footnotetextcopyrightpermission[1]{}
\begin{abstract}
Recent techniques on implicit geometry representation learning and neural rendering have shown promising results for 3D clothed human reconstruction from sparse video inputs.
However, it is still challenging to reconstruct detailed surface geometry and even more difficult to synthesize photorealistic novel views with animatated human poses.
In this work, we introduce PGAHum, a prior-guided geometry and appearance learning framework for high-fidelity animatable human reconstruction.
We thoroughly exploit 3D human priors in three key modules of PGAHum to achieve high-quality geometry reconstruction with intricate details and photorealistic view synthesis on unseen poses.
First, a prior-based implicit geometry representation of 3D human, which contains a delta SDF predicted by a tri-plane network and a base SDF derived from the prior SMPL model, is proposed to model the surface details and the body shape in a disentangled manner.
Second, we introduce a novel prior-guided sampling strategy that fully leverages the prior information of the human pose and body to sample the query points within or near the body surface.
By avoiding unnecessary learning in the empty 3D space, the neural rendering can recover more appearance details.
Last, we propose a novel iterative backward deformation strategy to progressively find the correspondence for the query point in observation space.
A skinning weights prediction model is learned based on the prior provided by the SMPL model to achieve the iterative backward LBS deformation.
Extensive quantitative and qualitative comparisons on various datasets are conducted and the results demonstrate the superiority of our framework. Ablation studies also verify the effectiveness of each scheme for geometry and appearance learning.
Codes will be released in \hyperlink{https://github.com/JLU-ICL/PGAHum}{https://github.com/JLU-ICL/PGAHum}.


\end{abstract}




\received{20 February 2007}
\received[revised]{12 March 2009}
\received[accepted]{5 June 2009}

\maketitle

\section{Introduction}

\begin{figure}[h]
  \includegraphics[width=\linewidth]{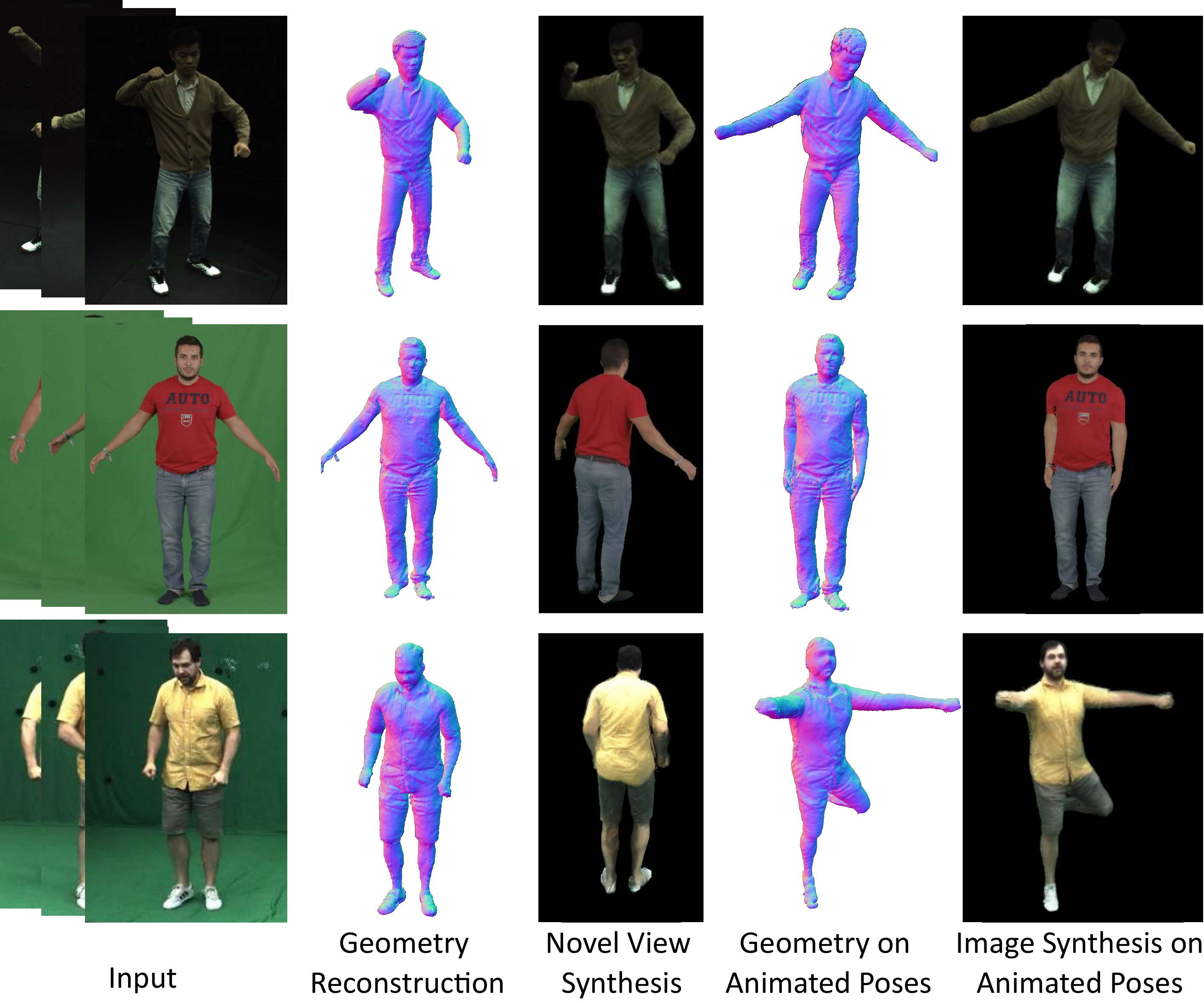}
  \caption{
  Given sparse input videos, our PGAHum can reconstruct high-fidelity animatable avatar with fine-grained geometry and appearance details on various datasets, e.g., ZJU-Mocap \cite{peng2021neural} (top), PeopleSnapshot \cite{alldieck2018detailed} (middle) and MonoCap \cite{peng2024animatable} (bottom). 
  }
  \label{fig:teaser}
\end{figure}


The digitization of the human body is crucial for various applications such as gaming, film, mixed reality, remote interaction and the metaverse.
In the industries, high-fidelity human body reconstruction typically requires acquiring data by multi-camera systems in well-equipped studios and building pre-captured templates with the assistance from skilled artists.
These requirements prohibit the conventional applications for consumers, such as personalized avatars used in AR/VR, body measurements, virtual try-on, etc. 


Human body reconstruction has been a popular and important research topic in recent years. 
Recent methods can reconstruct the clothed 3D human body from sparse videos, making the 3D human avatar acquisition more convenient and flexible. 
One series of clothing-aware body reconstruction methods use explicit mesh or fixed-resolution truncated signed distance fields (TSDFs) to represent the geometric shape of humans, while the textures are represented by vertex colors or UV maps. 
Based on existing statistical human models \cite{SMPL:2015, pavlakos2019expressive, osman2020star, jiang2020disentangled}, these methods \cite{ma2020learning, pavlakos2018learning, jiang2020bcnet} initialize 3D human avatar in some geometry representation of fixed-resolution, and then learn geometry offsets or displacements to enhance detailed surface geometry. 
While the spatiotemporal consistency of 3D human avatar can be obtained by building their human representation on top of models such as SMPL \cite{SMPL:2015}, the capability to express details is limited since the geometry offset learning is bound to a fixed number of geometric vertices.
Moreover, the appearance of the reconstructed human is not realistic enough.


With the success of neural implicit representations \cite{Park_2019_CVPR, mescheder2019occupancy, mildenhall2020nerf}, recent reconstruction methods combine neural implicit representations with neural rendering techniques and achieve promising results in geometry or appearance learning of 3D human. 
Some methods \cite{huang2020arch, zheng2021pamir, hong2021stereopifu, jiang2022selfrecon} aim to learn implicit representations of the human body which are capable of handling various topologies, adapting to different characters and clothing, as well as supporting animations with different poses. 
Meanwhile, a series of neural rendering methods \cite{jiang2022neuman, li2023posevocab, jayasundara2023flexnerf, liao2023high, zheng2022structured, xu2021h, weng2022humannerf, yu2023monohuman} mainly focus on synthesizing photorealistic novel viewpoint images, while not explicitly performing optimization for the geometry.
However, high-quality dense multi-view data may still be needed as the supervision for the neural rendering process.
On the other hand, there exist several works \cite{peng2021neural,ARAH:2022:ECCV,  peng2021animatable,peng2024animatable} 
simultaneously performs human geometry reconstruction and appearance learning.
While the reconstruction results capture the overall body shape, these methods struggle with finer geometry details, such as clothing wrinkles.
The main reason is the inherent non-rigid motion characteristics of humans and clothing may cause challenges for the correspondence searching during the implicit representation learning.
As the result of the sub-optimal geometry learning, the quality of view synthesis may also be affected to some extent.

In this work, we aim for high-fidelity animatable human reconstruction, i.e., high-quality geometry reconstruction with intricate human body and clothing details, as well as photorealistic image synthesis on novel views and unseen poses.
Comparing to previous works, we propose to more thoroughly exploit the priors from the 3D human models, e.g., SMPL \cite{SMPL:2015}, to enhance the geometry and appearance learning from the following aspects:
1) More fine-grained 3D human geometry representation, which is capable of expressing the surface geometric details as well as maintaining the spatiotemporal consistency for both the geometry and appearance reconstruction;
2) More human-centric implicit representation learning scheme, which is suitable for learning the 3D human models instead of treating the learning process similar to other objects or scenes;
3) Improved space transformation learning, which is important to learn a unified 3D human model in a canonical space with improved geometry and appearance details.
Specifically, we propose PGAHum, a novel framework which extensively utilizes 3D human priors to reconstruct high-fidelity animatable clothed humans from sparse videos.
The overall pipeline of PGAHum is similar to works \cite{ARAH:2022:ECCV,peng2024animatable} which simultaneously learn the geometry and appearance of the 3D human, we make improvements on three key modules to achieve reconstruction with more fine-grained geometry and appearance details.

First, we define a \textit{prior-based implicit geometry representation} to effectively learn the underlying body shape and surface details in a disentangled manner.
On top of the base signed distance field (SDF) derived from the SMPL model fitted to the subject, a tri-plane network is learned to predict a delta SDF layer for encoding the surface details.
With the strong prior from the SMPL model, the network can focus on learning the fine-grained geometry details rather than the overall body shape.
Second, to further confine the geometry and appearance learning on the fine details of the human body, a \textit{prior-guided sampling} scheme which fully leverages the prior information of the human pose and body is proposed.
Comparing to existing methods \cite{mildenhall2020nerf, wang2021neus} which mainly use the stratified sampling to sample query points in the full space, we first compute the ray-body intersection and only sample the points within or near the body surface, so that unnecessary sampling in empty 3D space can be avoided and the learned neural radiance filed (NeRF) can recover more appearance details.
Last, we propose an \textit{iterative backward deformation} strategy to warp query points in the observation space to the canonical space in a progressive manner, so that both the geometry and appearance of the unified human model can be better optimized.
Notably, we learn a backward skinning weights prediction model which takes the transformed query points to generate skinning weights for the iterative backward Linear Blend Skinning (LBS) deformation.
Comparing to the forward deformation \cite{ARAH:2022:ECCV}, our iterative backward deformation 
does not need to solve the root-finding problem to find the correspondances for the query points.
Also, it can effectively finds the appropriate correspondences through multiple backward deformation, alleviating the errors for the one-step backward deformation.

By combining above schemes, our PGAHum takes a solid step further to high-fidelity animatable human reconstruction, obtaining avatar with appealing geometry and appearance details for novel pose or view synthesis (see Figure \ref{fig:teaser}).
Extensive quantitative and qualitative comparisons are conducted and the results demonstrate the superiority of our framework.
Ablation studies also verify the effectiveness of each scheme for geometry and appearance learning.
In summary, our contributions are as follows:
\begin{itemize}
\item We propose PGAHum, a novel framework which extensively exploits the 3D human priors for high-fidelity animatable human reconstruction. Our results show more fine-grained geometry and appearances details than existing SOTA methods.
\item We disentangle the 3D clothed human with a prior-based implicit geometry representation.
Such fully implicit representation not only supports disentangled surface detail modeling, but also fits well for following NeRF learning.
\item We leverage the prior information of the human model and propose a novel prior-guided sampling strategy to avoid the unnecessary learning in the empty 3D space, so that the neural rendering can recover more appearance details.
\item We develop an iterative backward deformation strategy which reduces the computation cost for the forward deformation and the error for one-step backward deformation.
To achieve the iterative backward LBS deformation, a skinning weights prediction model is learned based on the prior provided by the SMPL model.
\end{itemize}


\section{Related Works}

\noindent{\textbf{Geometry-conditioned human reconstruction.}}
Based on existing statistical models \cite{SMPL:2015, pavlakos2019expressive, osman2020star, jiang2020disentangled}, some reconstruction methods \cite{bogo2016keep, omran2018neural, kanazawa2018end} use the prior human geometry and topology as the condition and learn to acquire model parameters from image inputs. 
To represent clothed human bodies, geometry offsets on top of the base geometry model are also learned \cite{ma2020learning,pavlakos2018learning, alldieck2018video} to express the clothing geometry. 
However, these representations mainly support compact clothing types and the reconstructed geometry is often coarse. 
Meanwhile, the clothing geometry can also be modeled separately \cite{jiang2020bcnet, ma2021scale} so that the base and surface geometry can be modeled in a more disentangled manner.
Additionally, to obtain more high-fidelity reconstruction, a series of works \cite{xu2018monoperfcap,habermann2019livecap,habermann2020deepcap} use a character-specific template to assist in pose tracking and performance capture. 
In general, these methods rely on explicit geometry templates or fixed-resolution representations, making it difficult to achieve fine-grained geometry and appearance reconstruction.

\noindent{\textbf{Implicit representation learning for human reconstruction.}}
%
Learning the neural implicit representation \cite{mescheder2019occupancy, wang2021neus} of 3D human  has been widely studied in recent years.
The PIFu series \cite{saito2019pifu,saito2020pifuhd,hong2021stereopifu} utilize pixel-aligned feature fusion to predict high-precision depth information for estimating the implicit function of  human body. 
PaMIR \cite{zheng2021pamir} combines the features obtained based on parametric body model and the image to learn an implicit function for human body. 
 ARCH \cite{huang2020arch} and ARCH++ \cite{he2021arch++} combine pixel-aligned features from semantic-aware geometric encoders and appearance encoders as inputs to learn a joint-space human body model represented as an occupancy field.
 These methods mainly train networks to predict the implicit values with the enhancement of the appearance features.
 Though certain level of geometry details can be reconstructed, they cannot learn the fine-grained geometry details by only learning one global implicit representation.

\noindent{\textbf{Neural rendering for human novel view synthesis.}}
Different from geometry reconstruction, neural rendering \cite{mildenhall2020nerf} has been explored for synthesizing novel views of human.
Also, methods \cite{jiang2020bcnet, jiang2022neuman, weng2022humannerf, yu2023monohuman, instant_nvr, jiang2023instantavatar} that support learning neural radiance field (NeRF) from dynamic animation sequences are also developed so that human in novel unseen poses can be rendered.
The key of these methods is to learn a deformation field or model to warp or transform the query point in the observation space to the canonical space, in which the radiance field can be optimized using input human with different poses.
However, these methods mainly focus on the appearance model learning and synthesizing realistic novel view images, and overlook the geometry model learning.
Without human geometry as guidance, the images synthesized at unseen poses may either contain artifacts or lack of details.

\noindent{\textbf{Geometry and appearance learning for human reconstruction.}}
Instead of only performing geometry reconstruction or appearance learning with NeRF, some of recent methods simultaneously learn models for both geometry and appearance for human reconstruction.
Neural Body \cite{peng2021neural} anchors a set of structured latent codes to the vertices of the SMPL mesh to represent an implicit human body and learns the implicit geometry and radiance fields from sparse videos. 
A-NeRF \cite{su2021nerf} estimates the 3D skeleton structure of a human body through pose estimation and performs skeleton-relative encoding for geometry and appearance learning.
The skeleton pose has also been used to define a pose-driven deformation field \cite{peng2021animatable,peng2024animatable} for the observation-to-canonical space transformation so that the geometry and color can be optimized in the canonical space. 
ARAH \cite{ARAH:2022:ECCV} performs simultaneous ray-surface intersection search and correspondence search to find the transformed query points in the canonical space using a joint root-finding algorithm.
Then, an SDF-based volume rendering is conducted to learn networks to predict the SDF and color values.
Comparing to these methods, we extensively use the human prior to enhance the geometry representation and neural rendering strategies for more fine-grained geometry and appearance learning.
\section{Method}

\begin{figure*}[h]
  \centering
  \includegraphics[width=\linewidth]{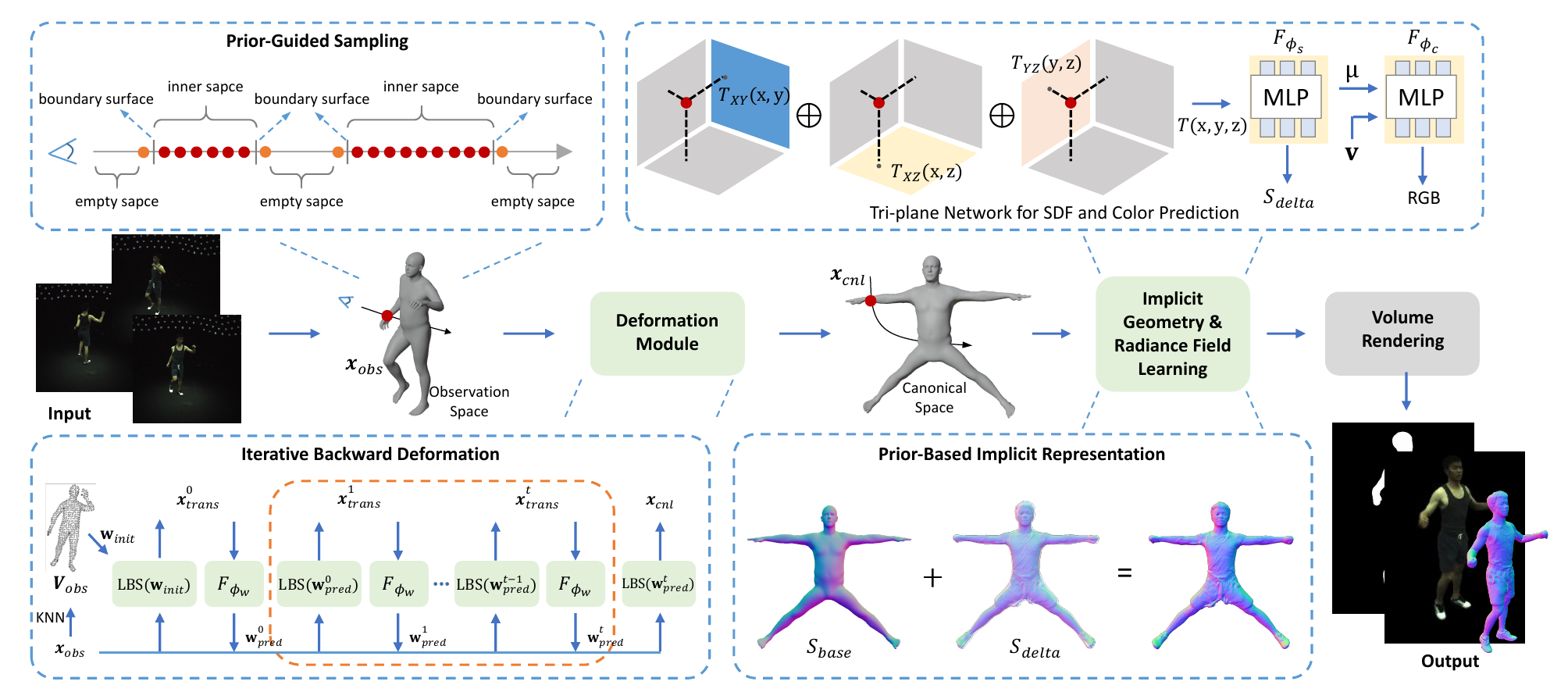}
  \caption{Overview of our pipeline. \rm{For an input view from the multi-view video frames with estimated human pose, we first utilize prior-guided sampling to sample points inside and around the human body based on the ray-body intersection, where the SMPL is used a prior for the body model. For a sampled point $\textbf{x}_{obs}$, we deform it to the corresponding point $\textbf{x}_{cnl}$ in a canonical space through the iterative backward deformation. 
  With the transformed points, we learn a prior-based implicit geometry representation which combines the prior SDF volume $\mathcal{S}_{base}$ derived from SMPL with $\mathcal{S}_{delta}$ predicted by a tri-plane network $F_{\phi_s}$ for modeling the human body with surface details in canonical space. 
  In addition, a feature vector $\mu$ produces from $F_{\phi_s}$ as well as view direction $\textbf{v}$ is passed to the color branch $F_{\phi_c}$ to get color value. 
  Finally, volume rendering is performed to render images, normal maps and subject mask for the loss computation.
  }
  }
  \label{figure:pipeline}
\end{figure*}

Our method aims to learn high-fidelity animatable human avatars from videos. The pipeline is shown in Figure~\ref{figure:pipeline}. Specifically, we first define a prior-based implicit geometry representation based on the SMPL model. It models the global body shape and local surface details in a fully implicit and disentangled manner  (Sec.~\ref{section:delta_sdf}). To learn this representation, given an input view with human pose, we design a prior-guided sampling scheme to sample 3D spatial points in observation space (Sec.~\ref{section:inner_sampling}). By leveraging the SMPL model, our sampling focuses more on spatial points within or near the human body. Then, we present an iterative backward deformation strategy to progressively warp the sampled points to the canonical space  (Sec.~\ref{section:iter_backward}). Finally, the warped points in the canonical space are used for volume rendering to compose the final rendered images (Sec.~\ref{section:rendering}). 
By imposing a series of loss functions on the rendered images and the geometry representation (Sec.~\ref{section:optimization}), our method learns a personalized animatable avatar that exhibits fine-grained surface geometry details and can be photorealisticlly rendered from different viewpoints and under novel poses.

\subsection{Preliminary}
\label{section:preliminary}
We adopt the SMPL model \cite{SMPL:2015} to provide strong priors for our key modules.
SMPL is a parametric representation of human bodies in a neutral pose and allows for easy manipulation using shape parameters \textbf{$\beta$} and pose parameters \textbf{$\theta$}. It typically defines 24 joints, including the hands, feet, head, torso, and other major body parts. The model also consists of a Linear Blend Skinning (LBS) model that deforms a template mesh to fit the shape of the human body, and a pose space model that parameterizes body pose using joint angles. It is generally used in body shape and pose estimation tasks. We utilize the prior bone matrix \textbf{B}, prior skinning weights $\Tilde{\textbf{W}} \in \mathbb{R}^{N \times J}$, prior signed distance field volume $\mathcal{S}_{base} \in \mathbb{R}^{256\times256\times256}$ of a canonical posed mesh, and vertices $\textbf{V}_{obs} \in \mathbb{R}^{N\times 3}$ of observation posed mesh for each subject in our method. $N$ and $J$ represents the number of vertices and joints. The skinning weights are used in the LBS algorithm to deform points in observation space to canonical space.

\subsection{Prior-Based Implicit Representation}
\label{section:delta_sdf}


SDF is a powerful implicit 3D shape representation that takes a 3D point as input and outputs the closest distance from that point to the object surface. Some methods~\cite{ARAH:2022:ECCV,jiang2022selfrecon} have represent and learn the overall geometry of human avatars using one global SDF. 
However, we observe that these methods cannot reconstruct fine geometry details. 
Hence, we propose a prior-based implicit geometry representation to effectively learn the underling body shape and surface details in a disentangled manner.

Specifically, our proposed representation $\mathcal{S}$ is defined in the canonical space with the star-pose. It consists of base geometry prior field $\mathcal{S}_{base}$ and geometry detail layer $\mathcal{S}_{delta}$, which effectively combines the advantages of the global body consistency derived from the prior SMPL model and the local detail modeling by the tri-plane representation~\cite{chan2022efficient}. The base geometry prior field $\mathcal{S}_{base}$ is represented by the SDF volume derived from the SMPL model fitted to each subject. Then, we use a neural network $F_{\phi_s}$ to learn the delta SDF $\mathcal{S}_{delta}$ for $\mathcal{S}_{base}$, where $\phi_s$ represents learnable parameters. More concretely, the $F_{\phi_s}$  contains a tri-plane representation $T = (T_{xy}, T_{yz}, T_{xz})$ and a shallow Multi-Layer Perceptron (MLP), where $T_{xy}$, $T_{yz}$, and $T_{xz}$ are three learnable feature planes that are orthogonal to each other and form a 3D cube of size $L^3$ centered at $(0,0,0)$. To learn the delta SDF $S_{delta}(\textbf{x})$ at position $\textbf{x}$, we project $\textbf{x}$ onto each of the three planes and query the corresponding features on each plane by bilinear interpolation. By concatenating these features and passing them into the MLP, we obtain $S_{delta}(\textbf{x})$.   
Finally, we combine both the $S_{base}$ and $S_{delta}$ to yield a complete SDF for any position $\textbf{x}$ in canonical space as:
\begin{equation}
\label{equation:delta_sdf}
    \mathcal{S}(\textbf{x}) = \mathcal{S}_{base}(\textbf{x}) + \mathcal{S}_{delta}(\textbf{x}).
\end{equation}

Based on this representation, when querying the SDF value for a point $\textbf{x}_{cnl}$ in canonical space, we first sample the $\mathcal{S}_{base}(\textbf{x}_{cnl})$ from SDF volume, and then predict the corresponding delta SDF value $\mathcal{S}_{delta}(\textbf{x}_{cnl})$ via $F_{\phi_s}$ to compute the final SDF value. Thanks to the strong SDF prior from $\mathcal{S}_{base}$, the network can focus on learning fine-grained geometry details on clothed human rather than the overall body shape.
In addition to the SDF value, a feature vector $\mu$ that is also output by $F_{\phi_s}$ is forwarded to a color branch to predict view-dependent RGB values. The color branch is also represented by a shallow MLP, denoted as $F_{\phi_c}$.


\subsection{Prior-Guided Sampling}
\label{section:inner_sampling}

Following~\cite{wang2021neus, ARAH:2022:ECCV}, we leverage the volume rendering technique to render images and supervise the SDF learning. In the vanilla volume rendering process \cite{mildenhall2020nerf, wang2021neus}, a stratified sampling strategy is used to sample spatial points on a ray, which introduces many unnecessary spatial point queries, hindering the efficiency and effectiveness of volume rendering on learning details. 
Considering the simple human body topology, accessible poses and easy-to-use SMPL model, we propose a prior-guided sampling strategy, which fully leverages the prior information of human poses and shapes.

Specifically, 
for a ray emitted from the camera center, we determine whether it will intersect with the human body shape which is represented by the SMPL model estimated at current viewpoint. 
We compute all intersection points with the human body and keep the depth value $z$ along the ray.
Since the camera center must be outside the human model, there must be an even number of intersection points, corresponding to a set of depth values $\{ z_0, z_1, \dots, z_{2n-1} \}$ in an ascending order, where the number of intersection points is $2n$. Each pair of intersection points with depth of $(z_{2i}, z_{2i+1})$ forms an intersection interval with length $l_i$, where $i=0, 1, \dots, n-1$.

To make our sampled points within and near the body surface, we extend the intersection interval to $(z_{2i}-0.1l_i, z_{2i+1}+0.1l_i)$. According to the length of each intersection interval, the sampling number of each interval is computed as: 
\begin{equation}
\label{equation:inner_sampling}
    Num_i = \frac{l_i}{\sum_{i=0}^{n-1 }{l_i}} * N_{samples},
\end{equation}
where $N_{samples}$ denotes the total number of sampling points required along a ray. Within each interval, $Num_i$ points will be uniformly sampled. If the ray does not intersect with the human body, we will uniformly sample points in the whole sampling span.

\subsection{Iterative Backward Deformation}
\label{section:iter_backward}

It is important to calculate accurate correspondences between the observation space and canonical space such that the appearance information in the observation space can be used to optimize the geometry and appearance representation in the canonical space. Backward deformation refers to transforming points from observation space to the canonical space, while forward deformation refers to the inverse progress. 
We follow \cite{jiang2023instantavatar,peng2024animatable} and adopt the backward deformation for more efficient deformation model learning.
However, the backward deformation in a single step is hard to guarantee accuracy. To address this issue, we propose an iterative backward deformation module so that the points in the observation space are progressively transformed to the appropriate positions in the canonical space through multiple iterations. 
This module will learn a skinning model $F_{\phi_w}$, a 4-layer MLP with trainable parameters $\phi_w$, in canonical space to predict skinning weights for LBS deformation.

Specifically, we first find the $K$ nearest neighbor vertices in the mesh $\textbf{V}_{obs}$ derived from the posed SMPL for each sampled point, and calculate the initial skinning weight for each sampled point as:
\begin{equation}
\label{equation:w_init}
    \textbf{w}_{init} = \sum_{k=1}^K \Tilde{\textbf{W}}(k) \cdot \frac{1}{\beta_k}, \quad \beta_k = \frac{d(k)}{\sum_{k=1}^K d(k)},
\end{equation}
where the $\Tilde{\textbf{W}}(k)$ denotes the prior skinning weights of the \textit{k-th} vertex, and $d(k)$ denotes the distance between the current sampled point and the \textit{k-th} nearest vertex. Afterwards, we deform the sampled point $\textbf{x}_{obs}$ to canonical space by backward LBS deformation:
\begin{equation}
\label{equation:lbs}
    \textbf{x}_{trans}^0 =  LBS(\textbf{x}_{obs}, \textbf{w}_{init}, \textbf{B}^{-1}) = (\sum_{i=1}^{J} \textbf{w}_{init}(i) \cdot \textbf{B}_i^{-1}) \textbf{x}_{obs},
\end{equation}
where $\textbf{x}_{trans}^0$ denotes the point in canonical space deformed from $\textbf{x}_{obs}$, $\textbf{B}=\{\textbf{B}_i\}^{24}$ denotes the bone-based transformation matrix which deforms from the canonical pose to the observation pose, and $\textbf{w}_{init}(i)$ is the weight corresponding to bone $\textbf{B}_i$. 

We then query skinning model $F_{\phi_w}$ with $\textbf{x}_{trans}^0$ to get predicted skinning weights $\textbf{w}_{pred}^0 = F_{\phi_w}(\textbf{x}_{trans}^0)$ for $\textbf{x}_{obs}$, and transform $\textbf{x}_{obs}$ to $\textbf{x}_{trans}^1$ using Equation~\eqref{equation:lbs} by $LBS(\textbf{x}_{obs}, \textbf{w}_{pred}^0, \textbf{B}^{-1})$.
However, the $\textbf{x}_{trans}^1$ maybe not be the accurate correspondence of $\textbf{x}_{obs}$ since the $\textbf{w}_{init}$ is computed from the prior SMPL model which only has an estimated base geometry. Hence, we additionally iterative $t$ times to get a more appropriate corresponding point in canonical space.
Finally, we obtain the corresponding point $\textbf{x}_{cnl}$ in canonical space for point $\textbf{x}_{obs}$.
The whole iterative backward deformation process is summarized in Algorithm \ref{algorithm:ibd}. 
We set $K=10$ and $t=3$ in our current implementation. 
\begin{algorithm}[t]
    \renewcommand{\algorithmicrequire}{\textbf{Input:}}
    \renewcommand{\algorithmicensure}{\textbf{Output:}}
    \caption{Iterative Backward Deformation }
    \label{algorithm:ibd}
    \begin{algorithmic}[1]
        \REQUIRE $\textbf{x}_{obs}, \textbf{V}_{obs}, \textbf{B}, \Tilde{\textbf{W}}, F_{\phi_w}$
        \STATE Initialize: $m \leftarrow 0$ 
        \STATE Find the $K$ nearest neighbor vertices to point $\textbf{x}_{obs}$ in $\textbf{V}_{obs}$ 
        \STATE Compute $\textbf{w}_{init}$ from $\Tilde{\textbf{W}}$ based on Equation ~\eqref{equation:w_init}
        \STATE $skinning\_weights \leftarrow \textbf{w}_{init}$ 
        \REPEAT
        \STATE $\textbf{x}_{trans}^m \leftarrow LBS(\textbf{x}_{obs}, skinning\_weights, \textbf{B}^{-1})$
        \STATE $skinning\_weights \leftarrow F_{\phi_w}(\textbf{x}_{trans}^m)$
        \STATE $m \leftarrow m+1$
        \UNTIL $m \geq t$
        \STATE $\textbf{x}_{cnl} \leftarrow LBS(\textbf{x}_{obs}, skinning\_weights, \textbf{B}^{-1})$
        \ENSURE Deformed point $\textbf{x}_{cnl}$ in canonical space
    \end{algorithmic}
\end{algorithm}

\subsection{Animatable Human Volume Rendering}
\label{section:rendering}
For the volume rendering, we follow \cite{wang2023pet} and incorporate the SDF values during the radiance field learning process.
Specifically, with the corresponding canonical points $\{\textbf{x}_{cnl}^i\}^{N_{samples}}$ for sampled points $\{\textbf{x}_{obs}^i\}^{N_{samples}}$ along the ray, we query $\mathcal{S}_{base}$ and $\mathcal{S}_{delta}$ to compute final SDF values by Equation~\eqref{equation:delta_sdf}. Besides, we query radiance field $F_{\phi_c}$ to get radiance (both radiance field and SDF are defined in canonical space). Finally, we accumulate the queried radiance $\{c_i\}^{N_{samples}}$ along the ray to get pixel color $C$ as:
\begin{equation}
\label{equation:rendering}
C = \sum_{i=1}^{N_{samples}} \prod _{j<i}(1-\alpha_j) c_i, \quad \alpha_i = 1 - \text{exp}(-\sigma_i \delta_i),
\end{equation}
\begin{equation}
\label{equation:sigmoid}
\sigma_i = s * (\Phi_s(\mathcal{S}(\textbf{x}_{cnl}^i))-1) \nabla \mathcal{S}(\textbf{x}_{cnl}^i) \cdot \textbf{v},
\end{equation}
where $\delta_i = ||\textbf{x}_{obs}^{i+1} - \textbf{x}_{obs}^i||_1 $ is the distance between two adjacent sampled points, $\Phi_s(\cdot)$ is the sigmoid function, $s \in \mathbb{R}$ is a learnable scale parameter, and $\textbf{v}$ is the view direction.

To further ensure the volume rendering to focus on the foreground region, we also synthesize a foreground mask based on the human shape represented by the SDF field.
Different from previous works \cite{peng2024animatable} which use the learned global SDF to obtain the mask, we directly use the prior base SDF for mask rendering to enhance the stability and facilitate the convergence of network learning.
In addition, we slightly inflate the mask with a certain distance to enable the learning of surface details.
Formally, the foreground mask is obtained by:
\begin{equation}
\label{equation:fg_mask}
    M(\textbf{x}) = 
    \begin{cases}
    0 & \mbox{if {$\mathcal{S}_{base}(\textbf{x}) - \tau$} > 0 } \\
    1 & \mbox{if {$\mathcal{S}_{base}(\textbf{x}) - \tau$} $\leq$ 0 }
    \end{cases},
\end{equation}
where $\tau$ (empirically set to 0.05) is a distance threshold for considering a point that is outside from the shape surface represented by $\mathcal{S}_{base}$ as foreground. 


\subsection{Optimization}
\label{section:optimization}

The final loss, including several photometric losses in observation space and multiple regularizers in canonical space, is defined as:
\begin{equation}
\label{equation:loss_total}
\begin{split}
    \mathcal{L}_{total} = &\lambda_1 \mathcal{L}_{rgb} + \lambda_2 \mathcal{L}_{lpips} + \lambda_3 \mathcal{L}_{nssim} + \\
    &\lambda_4 \mathcal{L}_{eikonal} + \lambda_5 \mathcal{L}_{skinning} + \lambda_6 \mathcal{L}_{mask}.
\end{split}
 \end{equation}

$\mathcal{L}_{rgb}$ is reconstruction loss by comparing the ground truth color $\Tilde{C}(r)$ and rendered color ${C}(r)$:
\begin{equation}
\label{equation:loss_rgb}
    \mathcal{L}_{rgb} = \sum_{r \in \mathcal{R}} || \Tilde{C}(r) - C(r)||_1,
\end{equation}
where $\mathcal{R}$ denotes the set of rays. 

$\mathcal{L}_{skinning}$ is loss for regularizing predicted skinning weights and prior ground truth:
\begin{equation}
    \mathcal{L}_{skinning} = || \textbf{w}_{pred}^t - \textbf{w}_{init} ||_1,
\end{equation}
where $\textbf{w}_{init}$ is sampled from $\Tilde{\textbf{W}}$ by $K$ nearest neighbors, and $\textbf{w}_{pred}^t$ is skinning weights queried from $F_{\phi_w}$ at the iteration step $t$. 

$\mathcal{L}_{lpips}$ is learned perceptual image patch similarity loss, $\mathcal{L}_{nssim}$ is similarity loss measured by Structural
Similarity Index Measure (SSIM), $\mathcal{L}_{eikonal}$ is Eikonal regularization \cite{icml2020_2086}, and $\mathcal{L}_{mask}$ is used to supervise the SDF with mask \cite{peng2024animatable}.
The coefficients $\lambda_1-\lambda_6$ are the weights of each loss function.
We set $\lambda_1=10$, $\lambda_2=\lambda_3=\lambda_5=\lambda_6=1$, $\lambda_4=0.1$.
After the first 50K iterations, $\lambda_5$ is set to 0.
\section{Experiments}

\subsection{Experimental Settings}

\textbf{Datasets.}
Following \cite{ARAH:2022:ECCV}, we use the \textbf{ZJU-MoCap} \cite{peng2021neural} dataset as our primary testbed. We also adopt the same setup with four cameras evenly spaced around the human subject and same training/test splits as \cite{ARAH:2022:ECCV} on this dataset.
We also conduct experiments on the \textbf{PeopleSnapshot} \cite{alldieck2018detailed} dataset, which contains monocular video of human subjects rotating in front of a camera. We follow the evaluation process implemented in \cite{jiang2023instantavatar}. 
Additionally, following the setting of \cite{peng2024animatable}, we conduct extra experiments on the \textbf{SyntheticHuman++} provided by \cite{peng2024animatable} to illustrate the quality of our reconstructed geometries. The dataset contains several sets of male and female animation sequences with both pose change and self-rotating motions. 

\noindent\textbf{Metrics.}
We evaluate our method on three tasks: novel view synthesis on training poses (NVS), generalization to unseen poses (Unseen), and geometry reconstruction (Recon). 
To evaluate the quality of synthesized images in the first two tasks, we use the PSNR, SSIM and LPIPS \cite{zhang2018unreasonable} metrics between rendered and corresponding ground-truth (GT) images. 
For geometry reconstruction, we evaluate our method and baselines on the training poses. 
As it is difficult to obtain GT geometry for dynamic humans, we follow ARAH \cite{ARAH:2022:ECCV} to generate pseudo-GT geometry for ZJU-MoCap by Pet-NeuS \cite{wang2023pet} and use Chamfer Distance (CD) to quantitatively evaluate geometry reconstruction.


\noindent\textbf{Implementation details.}
Our method is implemented with the PyTorch framework. The Adam \cite{kingma2014adam} is adopted for the training. The learning rate starts from $5e^{-4}$ and decays exponentially to $5e^{-5}$ along the optimization. The training is conducted on 4 A40 GPUs. Please refer to the supplementary for more details.


\subsection{Comparisons}

\begin{figure}[t]
  \centering
  \includegraphics[width=\linewidth]{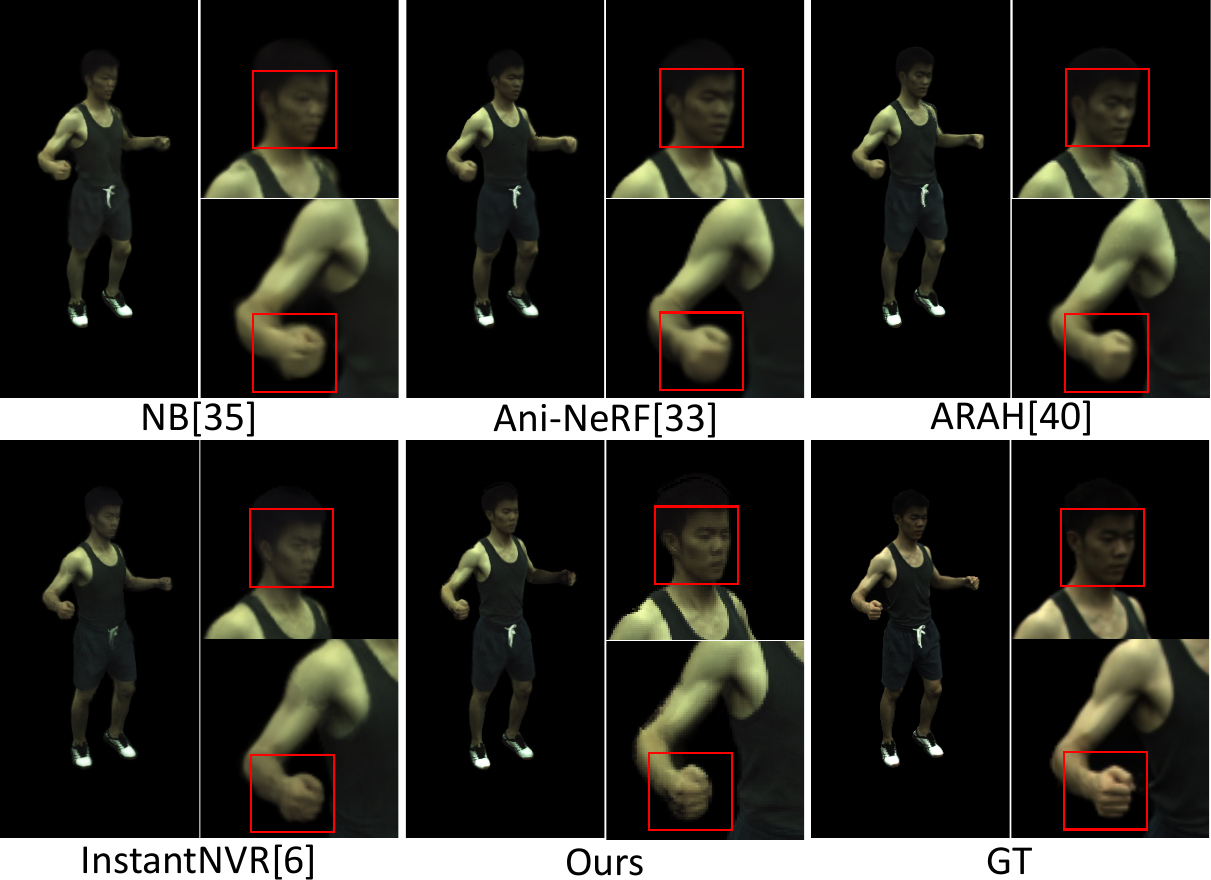}
  \caption{Qualitative results on ZJU-MoCap dataset for novel view synthesis on training poses.}
  \label{figure:nvs}
\end{figure}

\noindent\textbf{Novel view synthesis.}
In Table~\ref{tab:nvs} and Figure~\ref{figure:nvs}, we report the results of our method on novel view synthesis on ZJU-MoCap \cite{peng2021neural} dataset. In Figure~\ref{figure:nvs}, it can be observed that our method produces synthetic images with clearer human hands and heads, to the extent that even the number of fingers can be discerned. This is very difficult for other methods. As shown in Table~\ref{tab:nvs}, our approach demonstrates the best or comparable performance in terms of PSNR and LPIPS metrics in many cases compared to state-of-the-art methods. However, it shows inferior performance in terms of SSIM, possibly due to the limitations of our prior-guided sampling strategy in rendering the areas near the human contours, despite we can produce much better results within the contours. We will attempt to address this issue in our future works.

\begin{table*}
  \caption{Quantitative results for novel view synthesis on training poses. \rm{We compare PSNR ($\uparrow$), SSIM ($\uparrow$), and LPIPS ($\downarrow$) metrics on ZJU-MoCap dataset for novel view synthesis task. We bold the values with the \textbf{best} metric value and underline the \underline{second-best} ones.}}
  \label{tab:nvs}
  \resizebox{\linewidth}{!}{
  \begin{tabular}{c|ccc|ccc|ccc|ccc|ccc}
    \toprule
      & \multicolumn{3}{c|}{377} & \multicolumn{3}{c|}{387} & \multicolumn{3}{c|}{386} & \multicolumn{3}{c|}{393} & \multicolumn{3}{c}{394} \\
    \midrule
    Method & PSNR & SSIM & LPIPS & PSNR & SSIM & LPIPS & PSNR & SSIM & LPIPS & PSNR & SSIM & LPIPS & PSNR & SSIM & LPIPS \\
    \midrule
    NB~\cite{peng2021neural} & \textbf{28.1} & \textbf{0.956} & 0.080 & 26.7 & \underline{0.942} & 0.101 & \underline{29.0} & \underline{0.935} & 0.112 & \textbf{27.7} & \underline{0.939} & 0.105 & 28.7 & \underline{0.942} & 0.098 \\
    Ani-NeRF~\cite{peng2021animatable} & 24.2 & 0.925 & 0.124 & 25.4 & 0.926 & 0.131 & 25.6 & 0.878 & 0.199 & 26.1 & 0.916 & 0.151 & 27.5 & 0.924 & 0.142 \\
    A-NeRF~\cite{su2021nerf} & 27.2 & \underline{0.951} & 0.080 & 26.3 & 0.937 & 0.100 & 28.5 & 0.928 & 0.127 & \underline{26.8} & 0.931 & 0.113 & 28.1 & 0.936 & 0.103 \\
    ARAH~\cite{ARAH:2022:ECCV} & 27.8 & \textbf{0.956} & \underline{0.071} & \textbf{27.0} & \textbf{0.945} & \underline{0.079} &  \textbf{29.2} & \textbf{0.934} & \underline{0.105} & \textbf{27.7} & \textbf{0.940} & \underline{0.093} & \underline{28.9} & \textbf{0.945} & \underline{0.084} \\
    InstantNVR~\cite{instant_nvr} & 26.1 & 0.926 & 0.094 & 24.5 & 0.902 & 0.129 & 28.3 & 0.916 & 0.124 & 25.9 & 0.911 & 0.117 & 26.8 & 0.914 & 0.113 \\
    Ours & \underline{27.9} & 0.931 & \textbf{0.047} & \underline{26.9} & 0.918 & \textbf{0.066} & \textbf{29.2} & 0.893 & \textbf{0.083} & \textbf{27.7} & 0.910 & \textbf{0.074} & \textbf{29.0} & 0.919 & \textbf{0.063} \\
    
  \bottomrule
  \end{tabular}
  }
\end{table*}

\noindent\textbf{Generalization to unseen poses.}
On the ZJU-MoCap dataset, we typically train using the first 300 frames, while the frames after the 300th are used as unseen poses for testing generalization ability. 
In Table \ref{tab:unseen_zjumocap}, 
it can be observed that our method exhibits superior generalization performance on unseen poses. Similar to the above, although our method shows inferior performance in SSIM,  
it continues to maintain a lead in PSNR and LPIPS. 
In addition, the results in Table~\ref{tab:unseen_snapshot} demonstrate that our method achieves promising generalization performance on the PeopleSnapshot \cite{alldieck2018video} dataset as well. 
Please refer to the supplementary for more qualitative results.


\begin{table*}
  \caption{Quantitative results on ZJU-MoCap dataset for generalization to unseen poses.}
  \label{tab:unseen_zjumocap}
  \resizebox{\linewidth}{!}{
  \begin{tabular}{c|ccc|ccc|ccc|ccc|ccc}
    \toprule
      & \multicolumn{3}{c|}{377} & \multicolumn{3}{c|}{387} & \multicolumn{3}{c|}{386} & \multicolumn{3}{c|}{393} & \multicolumn{3}{c}{394} \\
    \midrule
    Method & PSNR & SSIM & LPIPS & PSNR & SSIM & LPIPS & PSNR & SSIM & LPIPS & PSNR & SSIM & LPIPS & PSNR & SSIM & LPIPS \\
    \midrule
    NB~\cite{peng2021neural} & 24.2 & \underline{0.917} & 0.119 & 22.7 & 0.902 & 0.135 & 26.1 & 0.894 & 0.171 & 23.5 & \underline{0.900} & 0.132 & 24.1 & \underline{0.888} & 0.150 \\
    Ani-NeRF~\cite{peng2021animatable} & 22.6 & 0.900 & 0.153 & 23.1 & 0.906 & 0.145 & 25.5 & 0.884 & 0.187 & 23.8 & 0.897 & 0.155 & 24.1 & 0.887 & 0.171 \\
    A-NeRF~\cite{su2021nerf} & 22.6 & 0.890 & 0.165 & 22.4 & 0.885 & 0.162 & 24.8 & 0.858 & 0.241 & 22.1 & 0.875 & 0.175 & 22.7 & 0.861 & 0.199 \\
    ARAH~\cite{ARAH:2022:ECCV} & \underline{25.5} & \textbf{0.933} & \underline{0.093} & \underline{24.2} & \textbf{0.917} & \underline{0.099} & \underline{27.0} & \textbf{0.910} & \underline{0.127} & \underline{24.4} & \textbf{0.915} & \underline{0.104} & \underline{25.2} & \textbf{0.908} & \underline{0.111} \\
    InstantNVR~\cite{instant_nvr} & 24.0 & 0.896 & 0.120 & 23.5 & 0.888 & 0.141 & 26.9 & 0.888 & 0.155 & 23.8 & 0.892 & 0.132 & 24.3 & 0.884 & 0.138 \\
    Ours & \textbf{25.6} & 0.904 & \textbf{0.070} & \textbf{25.6} & \underline{0.907} & \textbf{0.073} & \textbf{27.8} & \underline{0.895} & \textbf{0.092} & \textbf{25.4} & 0.891 & \textbf{0.079} & \textbf{26.0} & 0.866 & \textbf{0.086} \\
    
  \bottomrule
  \end{tabular}
  }
\end{table*}

\begin{table}
  \caption{Generalization to unseen poses on PeopleSnapshot dataset.}
  \label{tab:unseen_snapshot}
  \resizebox{\linewidth}{!}{
  \begin{tabular}{c|ccc|ccc}
    \toprule
    & \multicolumn{3}{c|}{male-3-casual} & \multicolumn{3}{c}{male-4-casual} \\
    \midrule
    Method & PSNR & SSIM & LPIPS & PSNR & SSIM & LPIPS\\
    \midrule
    NB~\cite{peng2021neural} & 24.94 & 0.9428 & 0.0326 & 24.71 & 0.9469 & 0.0423\\
    Anim-NeRF~\cite{chen2021animatable} & 29.37 & \underline{0.9703} & \textbf{0.0168} & \underline{28.37} & 0.9605 & \textbf{0.0268}\\
    InstantAvatar~\cite{jiang2023instantavatar} & \underline{29.65} & \textbf{0.9730} & \underline{0.0192} & 27.97 & \textbf{0.9649} & 0.0346\\
    Ours & \textbf{30.47} & 0.9478 & 0.0270 & \textbf{28.86} & \underline{0.9606} & \underline{0.0281}\\
    \midrule
    & \multicolumn{3}{c|}{female-3-casual} & \multicolumn{3}{c}{female-4-casual} \\
    \midrule
    Method & PSNR & SSIM & LPIPS & PSNR & SSIM & LPIPS\\
    \midrule
    NB~\cite{peng2021neural} & 23.87 & 0.9504 & 0.0346 & 24.37 & 0.9451 & 0.0382\\
    Anim-NeRF~\cite{chen2021animatable} & \underline{28.91} & \textbf{0.9743} & \textbf{0.0215} & 28.90 & 0.9678 & \textbf{0.0174}\\
    InstantAvatar~\cite{jiang2023instantavatar} & 27.90 & \underline{0.9722} & 0.0249 & \underline{28.92} & \textbf{0.9692} & \underline{0.0180}\\
    Ours & \textbf{30.86} & 0.9503 & 0.0358 & \textbf{32.49} & \underline{0.9689} & 0.0218\\
  \bottomrule
  \end{tabular}
  }
\end{table}

\begin{figure*}[h]
  \centering
  \includegraphics[width=0.8\linewidth]{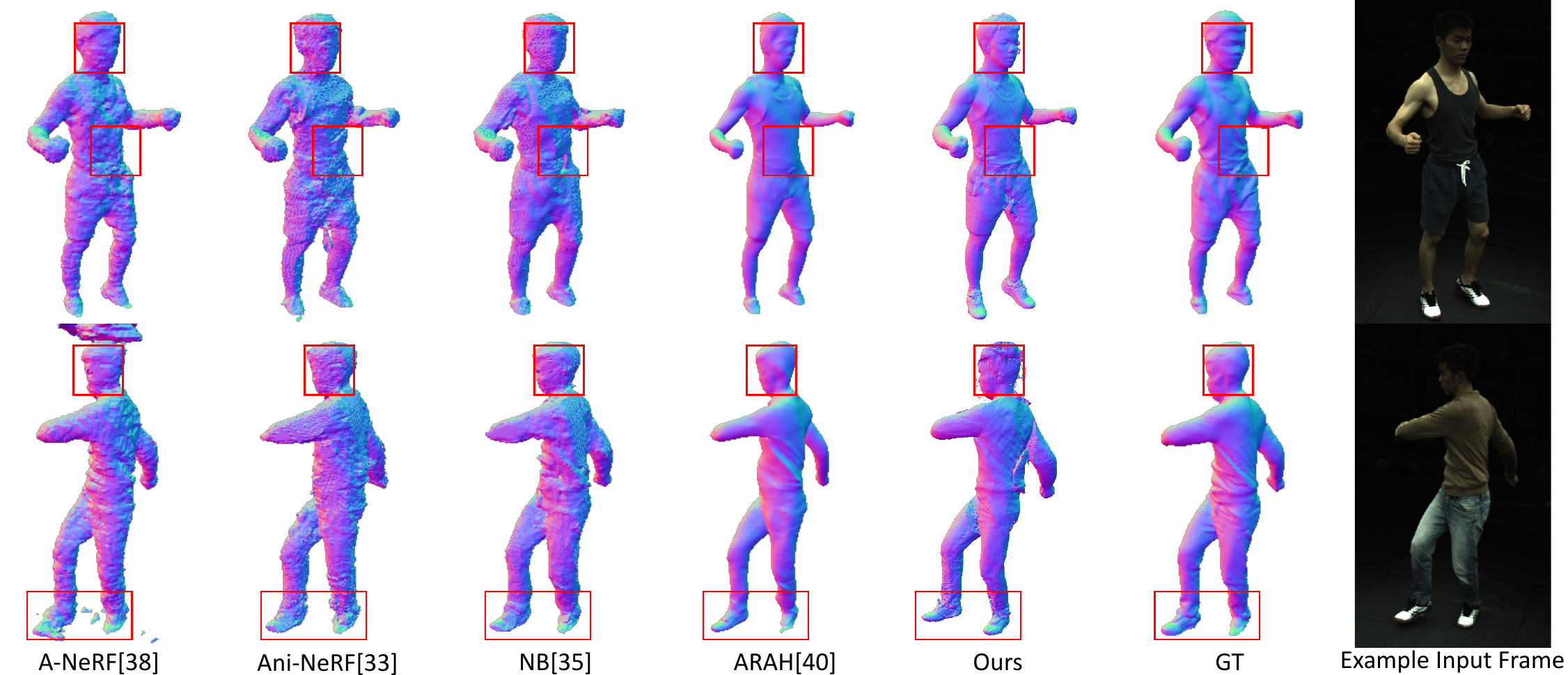}
  \caption{Qualitative results on ZJU-MoCap dataset for geometry reconstruction.}
  \label{figure:geometry_zjumocap}
\end{figure*}

\begin{figure}[h]
  \centering
  \includegraphics[width=\linewidth]{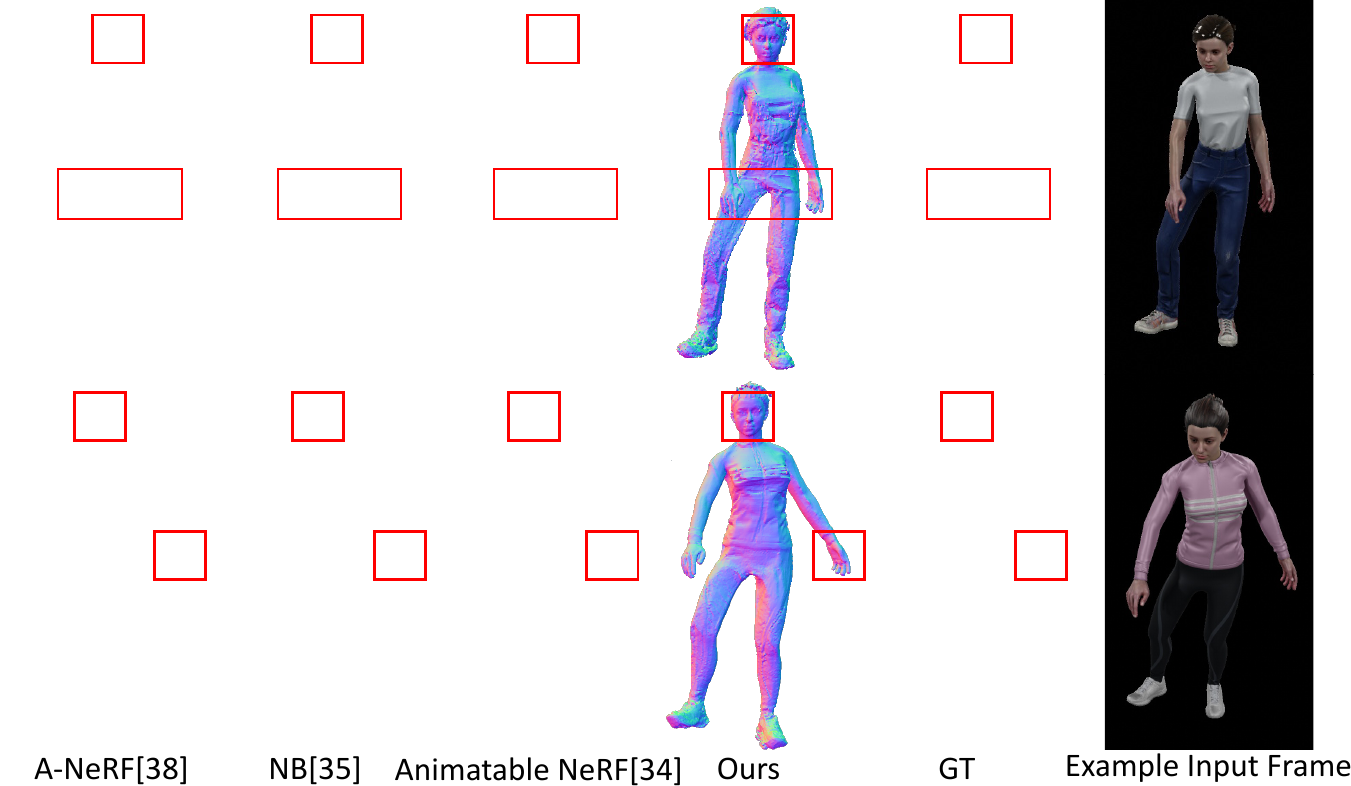}
  \caption{Qualitative results on SyntheticHuman++ dataset for geometry reconstruction.}
  \label{figure:geometry_synthetichuman}
\end{figure}

\noindent\textbf{Geometry reconstruction.}
In Figures \ref{figure:geometry_zjumocap} and  \ref{figure:geometry_synthetichuman}, we qualitatively compare our method with others \cite{peng2021neural, peng2021animatable, su2021nerf, peng2024animatable, ARAH:2022:ECCV} on ZJU-MoCap and SyntheticHuman++ dataset. It can be observed that our method reconstructs geometries with fine surface details. 
In Table \ref{tab:quantative_geometry}, we present a quantitative comparison of our method with NB \cite{peng2021neural} and ARAH \cite{ARAH:2022:ECCV} on the reconstructed geometries from the ZJU-MoCap dataset. It can be observed that our method consistently achieves the best results in terms of the Chamfer Distance (CD). 

\begin{table}
  \caption{Quantative comparison for geometry reconstruction in training poses. 
  \rm{We compare Chamfer Distance ($\downarrow$) metric on ZJU-MoCap dataset.}
  }
  \label{tab:quantative_geometry}
  \resizebox{\linewidth}{!}{
  \begin{tabular}{ccccccc}
    \toprule
    Subject & 377 & 386 & 387 & 393 & 394 & mean \\
    \midrule
    NB~\cite{peng2021neural} & 1.4417 & 1.3705 & 1.0814 & 1.4898 & 1.2034 & 1.3174 \\
    ARAH~\cite{ARAH:2022:ECCV} & 0.6846 & 0.2032 & 0.3168 & 0.8284 & 1.0341 & 0.6134 \\
    Ours & \textbf{0.6366} & \textbf{0.1842} & \textbf{0.3051} & \textbf{0.8009} & \textbf{1.0048} & \textbf{0.5863} \\
  \bottomrule
  \end{tabular}
  }
\end{table}

\subsection{Ablation Study}

We conduct ablation studies on subject 377 of ZJU-MoCap dataset to analyze the effectiveness of our proposed modules. 

\begin{figure}[h]
  \centering
  \includegraphics[width=\linewidth]{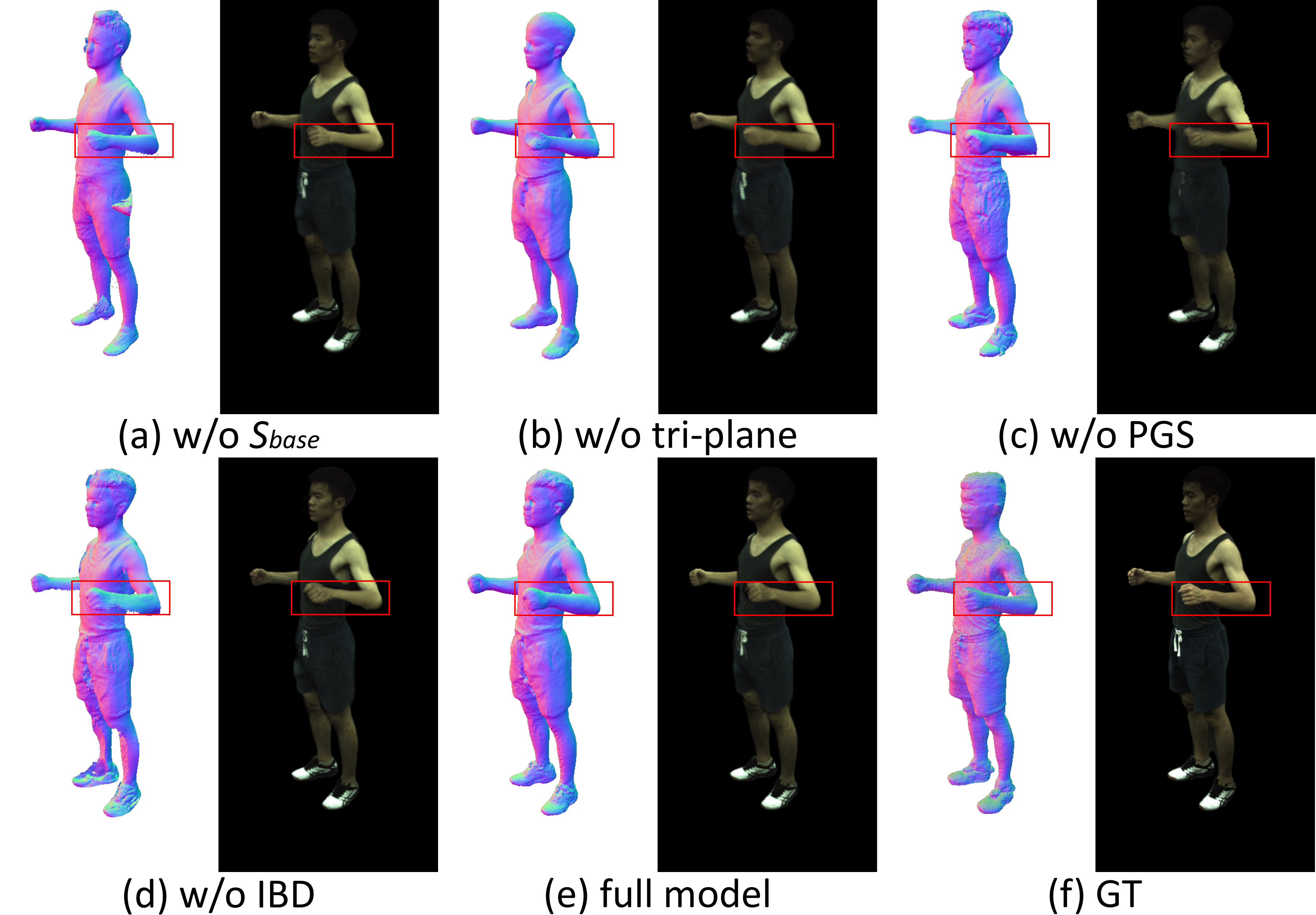}
  \caption{Qualitative results for ablation study.}
  \label{figure:ablation_study}
\end{figure}

\begin{table}
  \caption{Ablation study. \rm{IBD stands for iterative backward deformation and PGS stands for prior-guided sampling.}}
  \label{tab:ablation}
  \resizebox{\linewidth}{!}{
  \begin{tabular}{c|ccc|ccc|c}
    \toprule
    & \multicolumn{3}{c|}{NVS} & \multicolumn{3}{c|}{Unseen} & Recon \\
    \midrule
    Component & PSNR & SSIM & LPIPS & PSNR & SSIM & LPIPS & CD\\
    \midrule
    w/o IBD & 27.32 & 0.923 & 0.056 & 25.65 & 0.904 & 0.070 & \underline{0.6454}\\
    w/o tri-plane & 27.48 & 0.928 & \underline{0.051} & 25.64 & 0.904 & 0.069 & 0.6459\\
    w/o $\mathcal{S}_{base}$ & \underline{27.90} & \underline{0.931} & 0.057 & \textbf{25.92} & \underline{0.907} & \underline{0.063} & 0.6615\\
    w/o PGS & 27.10 & \textbf{0.935} & 0.055 & 25.54 & \underline{0.907} & \textbf{0.062} & 0.6570\\
    full model & \textbf{27.95} & \underline{0.931} & \textbf{0.047} & \underline{25.84} & \textbf{0.908} & \textbf{0.062} & \textbf{0.6366}\\
  \bottomrule
  \end{tabular}
  }
\end{table}

\noindent\textbf{Prior-based implicit representation.}
Our proposed method employs a combination of prior base geometry field $\mathcal{S}_{base}$ and geometry detail layer $\mathcal{S}_{delta}$ to represent an overall human SDF.
To validate its effectiveness, we conduct experiments by removing $\mathcal{S}_{base}$ module and solely utilizing a tri-plane network to learn the overall SDF. As shown in Table \ref{tab:ablation}, when $\mathcal{S}_{base}$ is dropped (w/o $\mathcal{S}_{base}$), there is a certain degree of performance decline in PSNR, LPIPS, and SSIM for both novel view synthesis and generalization to unseen poses. 
In Figure \ref{figure:ablation_study}(a), we can also observe that without this module, some ghosting artifacts appear at the end of the left elbow and on the pants in the synthesized images. We attribute this to the lack of underlying models to support the neural network in learning sufficiently robust and fine-grained representations for certain areas which are textureless or near the boundary.

In our method, we use a tri-plane representation to model $\mathcal{S}_{delta}$. Similarly, we conduct an ablation experiment by replacing it with a naive MLP network. In the row w/o tri-plane of Table~\ref{tab:ablation} and Figure~\ref{figure:ablation_study}(b), it can be seen that, 
due to the powerful expressive capability of the tri-plane representation, our method is able to achieve improved geometry reconstruction and novel view synthesis.

\noindent\textbf{Prior-guided sampling.}
To validate the effectiveness of our proposed prior-guided sampling strategy, we conduct experiments by replacing this module with naive uniform sampling. As observed in Table \ref{tab:ablation}, upon removing this module, there is a certain degree of performance decline in PSNR, LPIPS, and SSIM for both NVS and Unseen tasks. In Figure \ref{figure:ablation_study}(c), we notice obvious blurring in the arms and facial regions of the synthesized images. 
This is because the uniform sampling distracts the model's focus on learning the representative regions of the human body and introduces unnecessary points in empty space, which may lead to degraded performance on geometry and appearance learning.

\noindent\textbf{Iterative backward deformation.}
To investigate the effectiveness of our proposed iterative backward deformation, we conduct an ablation experiment by replacing this module with one-step backward skinning  deformation module. 
As observed in the row w/o IBD of Table \ref{tab:ablation}, without this module, there is a decrease of performance in PSNR, SSIM, and LPIPS for all tasks. In Figure \ref{figure:ablation_study}(d), we notice blurry patches and noise in the hand and pants regions of the subject. These results verify the iterative backward deformation module can effectively transform points from the observation space to the canonical space.
By performing multiple deformations, the transformed points can gradually converge to the actual corresponding canonical points, alleviating the errors by performing the single-step deformation.


\section{Conclusion}

We introduce PGAHum, a novel framework that takes a solid step further to high-fidelity animatable human reconstruction from multi-view or monocular videos.
To achieve fine-grained geometry and appearance learning, we effectively use human priors in three novel modules.
First, the prior-based implicit geometry representation combines the advantages of the global body consistency derived from the prior SMPL model and the powerful local detail modeling by the tri-plane representation, allowing the network to focus on learning fine surface details of clothed human.
Second, the prior-guided sampling leverages prior human pose and shape information to constrain sampling around human body, which encourages the volume rendering to learn more appearance details.
Last, the iterative backward deformation warps the query points using the skinning weight model learned based on the initial SMPL weights, ensuring efficient and accurate space transformation to facilitate the optimization in the canonical space.


Despite our PGAHum achieves promising results with intricate geometry details such as clothing wrinkles and more photorealistic novel view synthesis for human in unseen poses, it still faces some limitations.
First, due to the more computation cost on the ray-body intersection for the prior-guided sampling and the tri-plane network optimization for learning the disentangled geometry representation, the whole pipeline takes a relatively long training time. 
Next, similar to existing works \cite{ARAH:2022:ECCV,peng2024animatable}, our method may still not work well for reconstructing human with loose clothing, as our point sampling is constrained to the inner or near-body region for improved body surface learning.
Also, the geometry representation and the point deformation model learning may also be challenging if the points sampled for lose clothing are too far from the human body.
How to improve the computation efficiency of our prior-guided framework and generalize it to reconstruct human with lose clothing are interesting directions for future works.

\bibliographystyle{ACM-Reference-Format}
\bibliography{sample-authordraft}

\section{Supplementary}

In this supplementary material, we first introduce some additional implementation details in Section~\ref{section:prior_data_preprocessing}, including how the prior data is preprocessed, some specific implementation details of our method, and the reproduction details of the baseline methods. Then, we present more experimental results in Section~\ref{section:nvs}, including the three main tasks our method focuses on: novel view synthesis, generalization to unseen poses, and geometry reconstruction.

\subsection{More Implementation Details}

\noindent\textbf{Prior data preprocessing.}
\label{section:prior_data_preprocessing}
We use SMPL model provided by the datasets, such as ZJU-MoCap and MonoCap, as prior for prior-guided sampling and iterative backward deformation.
Specifically, given the SMPL shape and pose parameters in the observation space, we aim to obtain a dilated mesh so that when performing the prior-guided sampling, more points could be sampled near the boundary region.
To achieve goal, we first obtain the SMPL model mesh and convert it into a Signed Distance Function (SDF) representation.
Then, from this SDF, we extract the level set with distance equal to 0.05 meters and perform Marching Cube to obtain a dilated mesh corresponding to this level set.
Meanwhile, from the SMPL shape and pose parameters (star-pose) in the canonical space, we obtain the prior SDF volume field as $\mathcal{S}_{base} \in \mathbb{R}^{256 \times 256 \times 256}$ to represent the base geometry field.
Additionally, we maintain the prior skinning weights $\Tilde{\textbf{W}}$ from the default SMPL model, which will be utilized in the iterative backward deformation module.


\noindent\textbf{Implementation details.}
\label{section:implementation_details}
Both the $F_{\phi_s}$ and $F_{\phi_c}$ contain a 3-layer MLP. The Adam optimizer is adopted for training. For the skinning weights prediction model $F_{\phi_w}$, we set the initial learning rate to 1e-4, while for the SDF model $F_{\phi_s}$ and $F_{\phi_c}$, we set the initial learning rate to 5e-4. All of these models utilize the exponential learning rate decay method, where the learning rate gradually decreases exponentially as the global step increases.
We train our models with a batch size of 1, using 4 patches of $32 \times 32$ in height and width dimensions, resulting in 4096 rays per batch. With the patch sampling method, we can introduce $\mathcal{L}_{lpips}$ and $\mathcal{L}_{nssim}$ as supervision terms.
As mentioned in the main paper, we cast the rays and obtain the intersection points with the prior posed mesh to define the intersection interval bounded by two depth values along the ray. We then sample 154 ($ = \lceil 128*(1+0.2) \rceil$) points on each ray. Our model is trained for approximately 200-300K iterations, depending on the amount of training data, and takes about 3 days on 4 NVIDIA A40 GPUs for one subject. For comparison, it takes about 1.5 days for ARAH \cite{ARAH:2022:ECCV} to train 80K iterations for one subject on 4 NVIDIA 2080Ti GPUs. All the train-test splits follow the methods we compared.


\noindent\textbf{Implementation details for baselines.}
\label{section:implementation_details_for_baselines}
For the quantitative evaluation on the geometry reconstruction task in Table 4 of Neural Body \cite{peng2021neural} and ARAH \cite{ARAH:2022:ECCV}, we use the official code and provided pretrained weights without modification to obtain the first view and first frame posed mesh. For the quantitative evaluation on novel view synthesis and generalization to unseen pose in Tables 1, 2, and 3, we use the reported data in \cite{ARAH:2022:ECCV} and \cite{jiang2023instantavatar}.
It's worthy noting that when comparing with the InstantNVR \cite{instant_nvr} method, we directly use their official code and train on the original ZJU-MoCap dataset \cite{peng2021neural} for a fair comparison. Additionally, when computing the PSNR, SSIM, and LIPIS metrics, we use the bounding box mask to crop out the main human region from the original images. This is the reason why their results in Tables 1 and 2 are slightly inferior than those reported in their papers.

\begin{figure}[t]
  \centering
  \includegraphics[width=\linewidth]{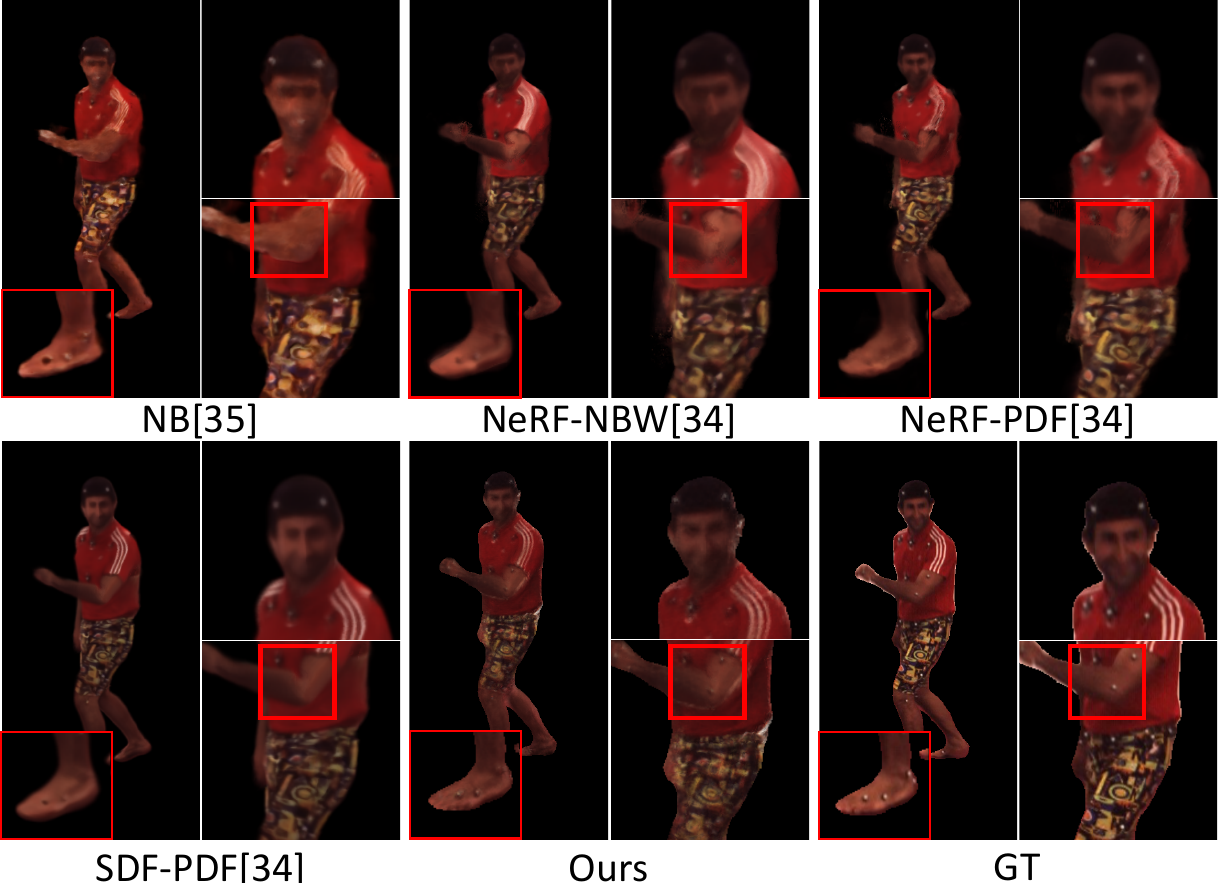}
  \caption{Additional qualitative comparison results for novel pose synthesis on H3.6M dataset.}
  \label{figure:unseen_pose_h36m}
\end{figure}

\begin{figure*}[t]
  \centering
  \includegraphics[width=\linewidth]{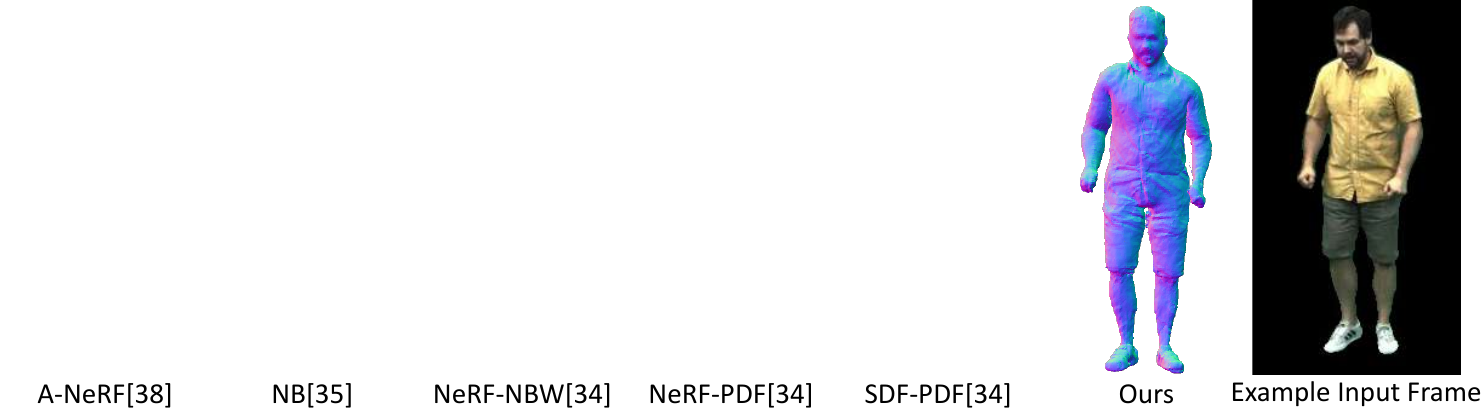}
  \caption{Additional qualitative comparison results for geometry reconstruction on MonoCap dataset.}
  \label{figure:geometry_monocap}
\end{figure*}

\subsection{More Experimental Results}

\begin{figure*}[t]
  \centering
  \includegraphics[width=0.85\linewidth]{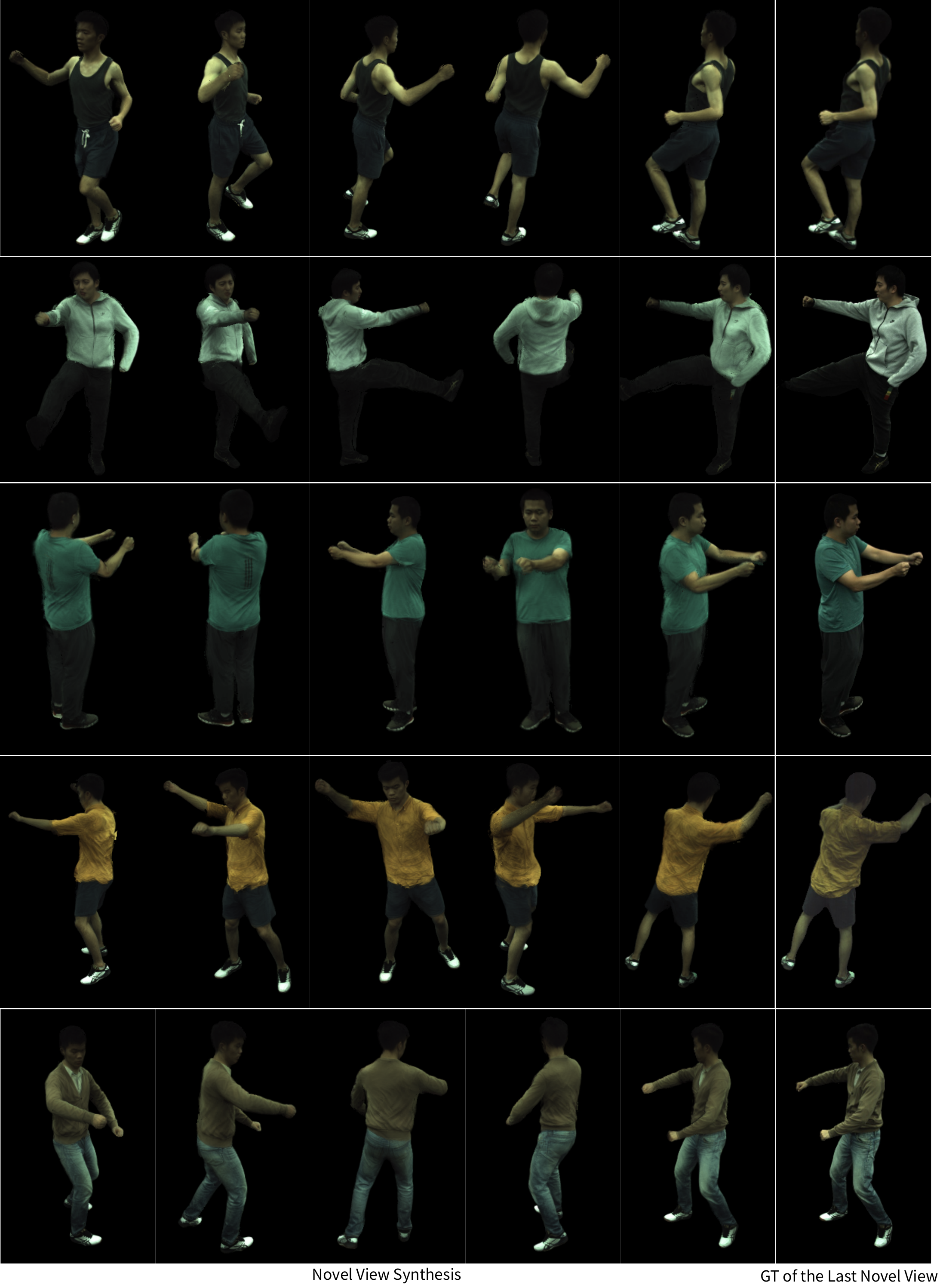}
  \caption{Additional qualitative results for novel view synthesis on ZJU-MoCap training poses.}
  \label{figure:nvs_zjumocap}
\end{figure*}

\noindent\textbf{Novel view synthesis on training poses.}
\label{section:nvs}
The proposed method can synthesize images from various novel viewpoints, allowing us to carefully observe the human pose. In Figure~\ref{figure:nvs_zjumocap}, we present additional qualitative results for novel view synthesis on the ZJU-MoCap training poses.
The figure illustrates the rendered images of the posed human from multiple views in the first five columns, and the last column shows the corresponding ground truth of the last novel view.

\begin{figure*}[t]
  \centering
  \includegraphics[width=0.8\linewidth]{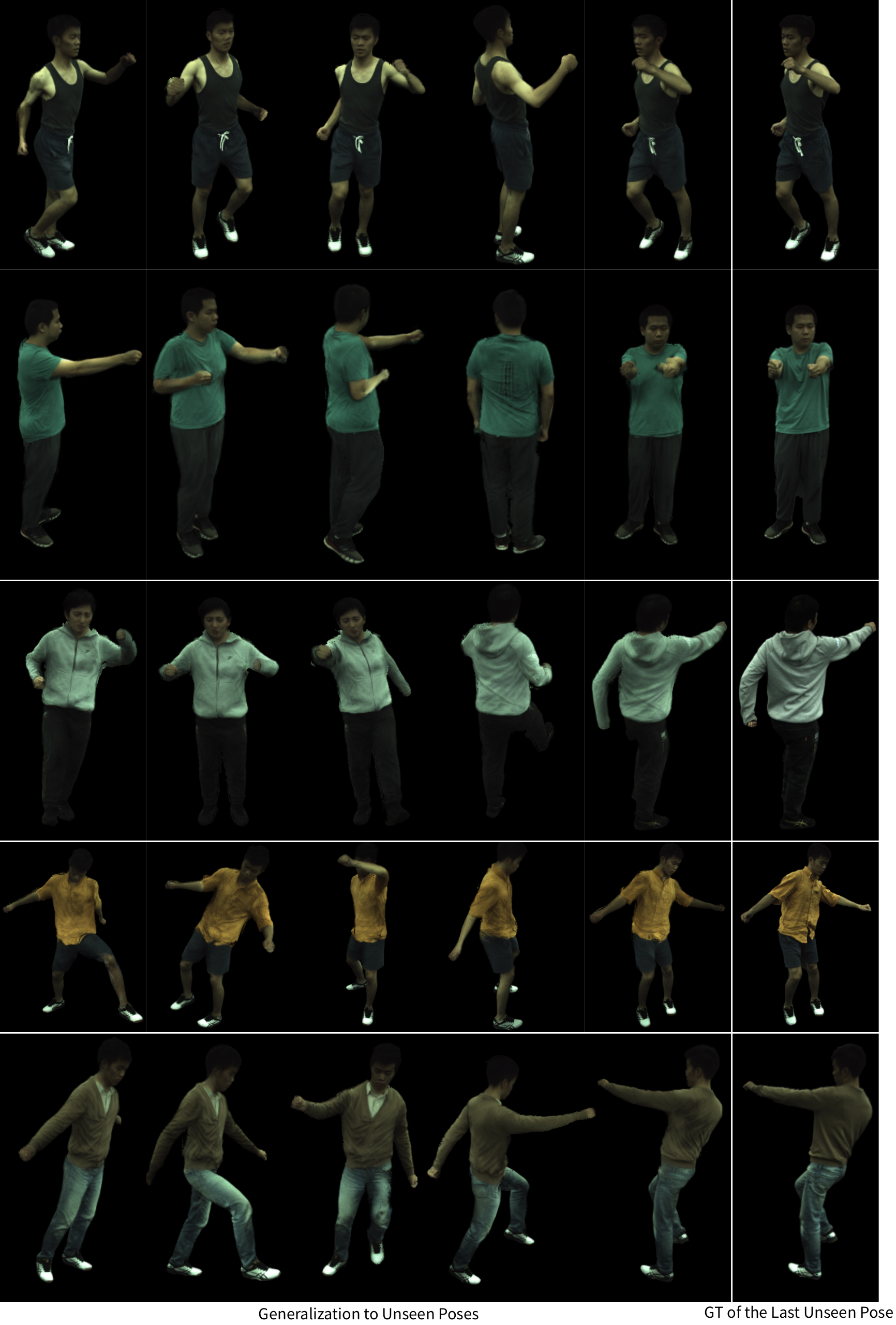}
  \caption{Additional qualitative results for novel pose synthesis on ZJU-MoCap unseen poses.}
  \label{figure:unseen_zjumocap}
\end{figure*}

\noindent\textbf{Generalization to unseen poses.}
\label{section:unseen}
The proposed method can synthesize images from various novelviewpoints and human poses.
In Figure~\ref{figure:unseen_pose_h36m}, we present additional qualitative comparison results with \cite{peng2024animatable, peng2021neural} for unseen pose synthesis on the H3.6M dataset \cite{peng2024animatable}. It can be observed that the feet and elbow parts rendered by our method are clearer than those of other approaches, such as NeRF-NBW, NeRF-PDF, and SDF-PDF, which were reported in \cite{peng2024animatable}.
Furthermore, we showcase more qualitative results for novel pose synthesis on the ZJU-MoCap training poses in Figure~\ref{figure:unseen_zjumocap}, i.e., images of different unseen poses are rendered from the same viewpoint.

\begin{figure*}[t]
  \centering
  \includegraphics[width=\linewidth]{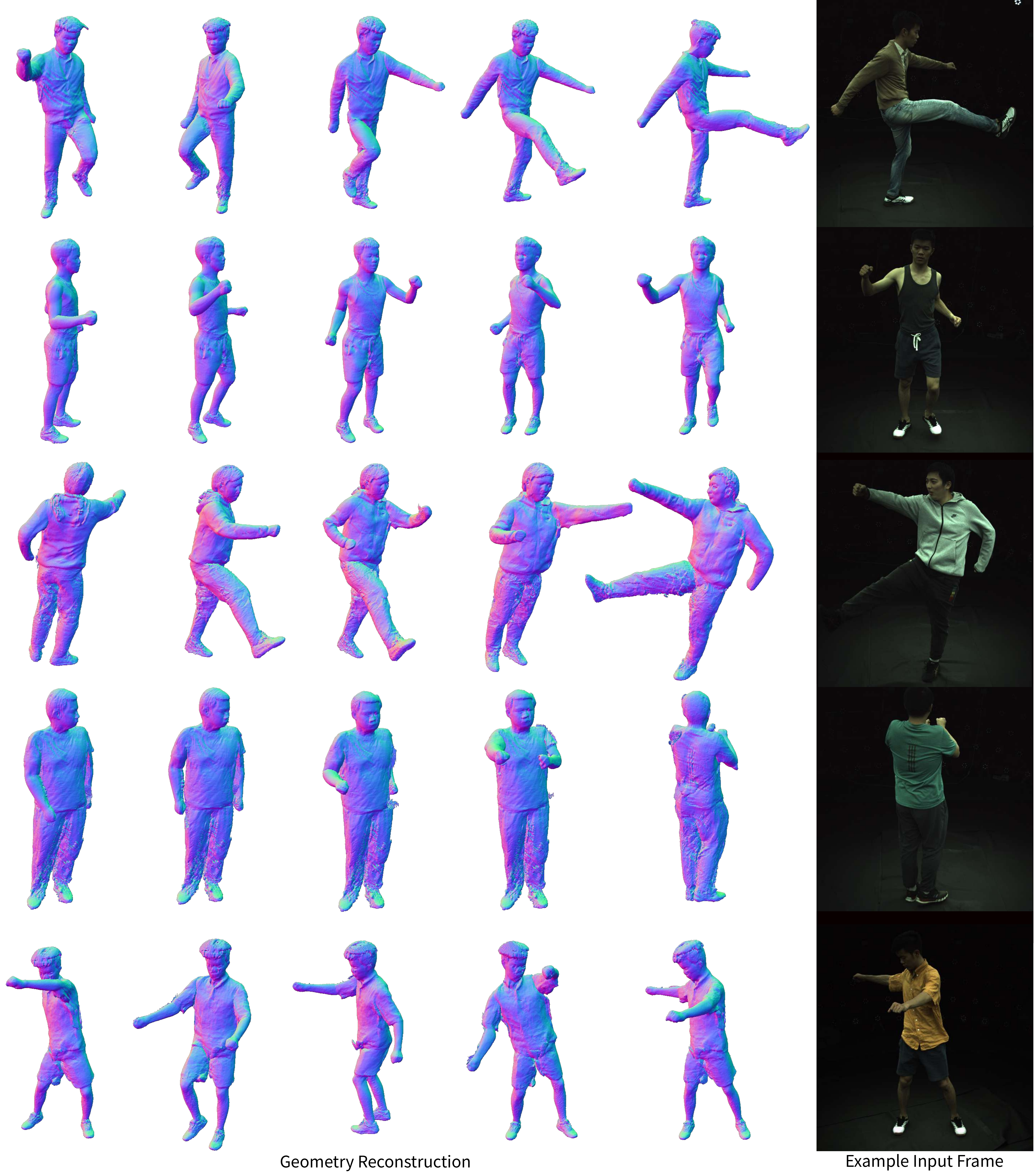}
  \caption{Additional geometry reconstruction on ZJU-MoCap datasets.}
  \label{figure:zjumocap_unseenpose_normal}
\end{figure*}
\begin{figure*}[t]
  \centering
  \includegraphics[width=\linewidth]{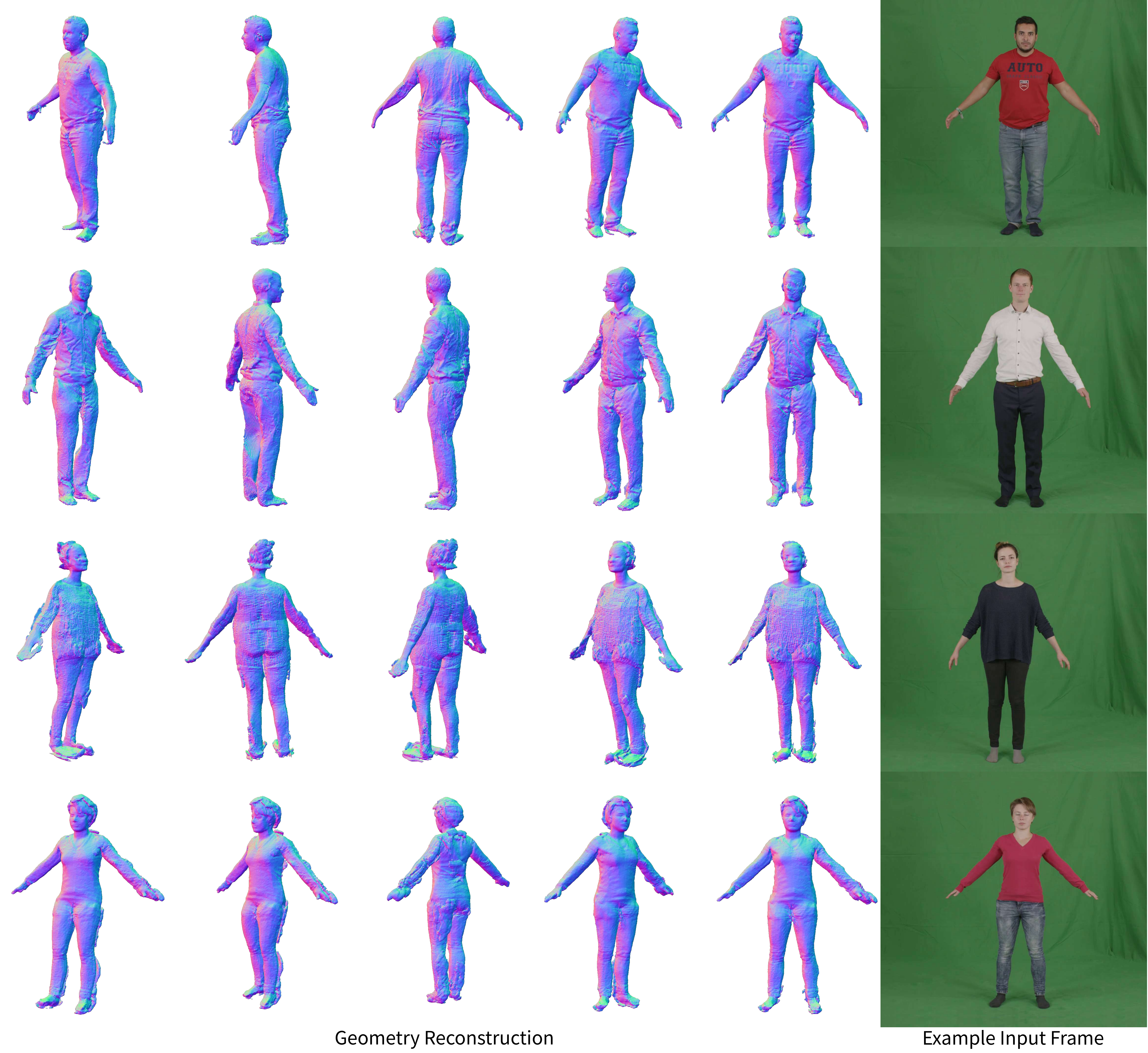}
  \caption{Additional geometry reconstruction on PeopleSnapshot datasets.}
  \label{figure:peoplesnapshot_unseenpose_normal}
\end{figure*}
\begin{figure*}[t]
  \centering
  \includegraphics[width=\linewidth]{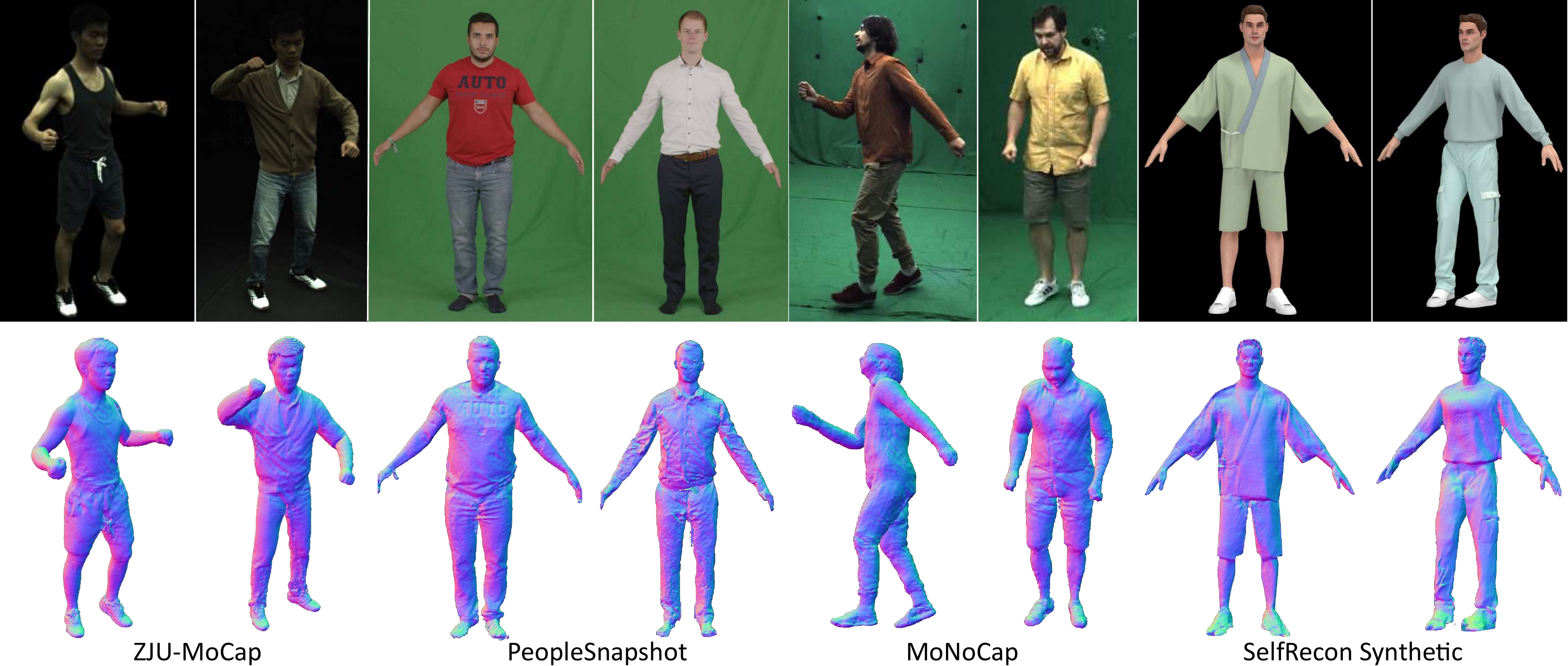}
  \caption{Additional geometry reconstruction on more datasets.}
  \label{figure:geometry_more_datasets}
\end{figure*}

\noindent\textbf{Geometry reconstruction.}
\label{section:geometry}
The proposed method can reconstruct geometry in various human poses with fine geometric details.
In Figure~\ref{figure:geometry_monocap}, we present additional qualitative comparison results with \cite{peng2024animatable, peng2021neural, su2021nerf} for geometry reconstruction on the MonoCap dataset. We can observe that the geometry reconstructed by our method exhibits superior performance in terms of geometric details, such as the folds of the shirt and trousers.
Furthermore, we show more qualitative results for geometry reconstruction on the ZJU-MoCap datasets in Figure~\ref{figure:zjumocap_unseenpose_normal} and Figure~\ref{figure:peoplesnapshot_unseenpose_normal}. Additionally, we illustrate more qualitative results for geometry reconstruction on additional datasets, such as MonoCap~\cite{peng2024animatable} and SelfRecon Synthetic~\cite{jiang2022selfrecon}, in Figure~\ref{figure:geometry_more_datasets}.


%

\end{document}